\documentclass[10pt,twocolumn,letterpaper]{article}

\usepackage{cvpr}              %

\usepackage{graphicx}
\usepackage{amsmath}
\usepackage{amssymb}
\usepackage{booktabs}
\usepackage[accsupp]{axessibility}  %
\usepackage{multirow}
\usepackage{adjustbox}
\usepackage[table,xcdraw]{xcolor}
\usepackage[shortlabels]{enumitem}

\usepackage[pagebackref,breaklinks,colorlinks]{hyperref}

\usepackage[capitalize]{cleveref}
\crefname{section}{Sec.}{Secs.}
\Crefname{section}{Section}{Sections}
\Crefname{table}{Table}{Tables}
\crefname{table}{Tab.}{Tabs.}

\def\cvprPaperID{23}
\def\confName{-}
\def\confYear{-}

\begin{document}

\title{GeoMultiTaskNet: remote sensing unsupervised domain adaptation using geographical coordinates} 

\author{Valerio Marsocci\\
Sapienza University of Rome\\
{\tt\small valerio.marsocci@uniroma1.it}
\and
Nicolas Gonthier\\
Univ Gustave Eiffel, IGN, ENSG, LASTIG\\
{\tt\small nicolas.gonthier@ign.fr}
\and
Anatol Garioud\\
IGN\\
{\tt\small anatol.garioud@ign.fr}
\and
Simone Scardapane\\
Sapienza University of Rome\\
{\tt\small simone.scardapane@uniroma1.it}
\and
Clément Mallet\\
Univ Gustave Eiffel, IGN, ENSG, LASTIG\\
{\tt\small clement.mallet@ign.fr}
}
\maketitle

\begin{abstract}
Land cover maps are a pivotal element in a wide range of Earth Observation (EO) applications. However, annotating large datasets to develop supervised systems for remote sensing (RS) semantic segmentation is costly and time-consuming. Unsupervised Domain Adaption (UDA) could tackle these issues by adapting a model trained on a source domain, where labels are available, to a target domain, without annotations. UDA, while gaining importance in computer vision, is still under-investigated in RS. Thus, we propose a new lightweight model, GeoMultiTaskNet, based on two contributions: a GeoMultiTask module (GeoMT), which utilizes geographical coordinates to align the source and target domains, and a Dynamic Class Sampling (DCS) strategy, to adapt the semantic segmentation loss to the frequency of classes. This approach is the first to use geographical metadata for UDA in semantic segmentation. It reaches state-of-the-art performances (47,22\% mIoU), reducing at the same time the number of parameters (33M), on a subset of the FLAIR dataset, a recently proposed dataset properly shaped for RS UDA, used for the first time ever for research scopes here. 
\end{abstract}

\section{Introduction}
\label{sec:intro}

Accurate land cover information is crucial for a wide range of applications, including environmental monitoring and management \cite{9705087, 9127795, kemker2018algorithms}, urban planning, and monitoring \cite{shelestov2021extension,buhler2021application}. In particular, semantic segmentation is a key task in the analysis of very high-resolution (VHR) remote sensing (RS) images, as it enables the automatic categorization of land cover \cite{marsocci2021mare}. However, annotating large datasets for supervised learning is costly and time-consuming, especially when not all data are acquired contemporaneously \cite{cbt}. 

In this context, unsupervised domain adaptation (UDA) offers a promising solution for adapting a model trained on a source domain to a target domain, without the need for annotations \cite{ganin2015unsupervised, kang2019contrastive, long2016unsupervised}, reducing domain shift. Although this task is gaining importance in computer vision (CV) \cite{Hoyer_2022_CVPR, hoyer2022hrda, xie2022sepico}, in RS it is still under-investigated. 
On one hand, often new RS UDA methods are applied on datasets not properly developed for this purpose \cite{rottensteiner2012isprs} and, consequently, far from the real-world UDA scenario. On the other hand, general CV models are often applied to RS images, with little regard to the EO peculiarities. A clear example is the use of metadata, such as geographical coordinates, which are often discarded\cite{salcedo2020machine, zhu2017deep}. 

\begin{figure}
  \centering
  \includegraphics[width=0.75\linewidth]{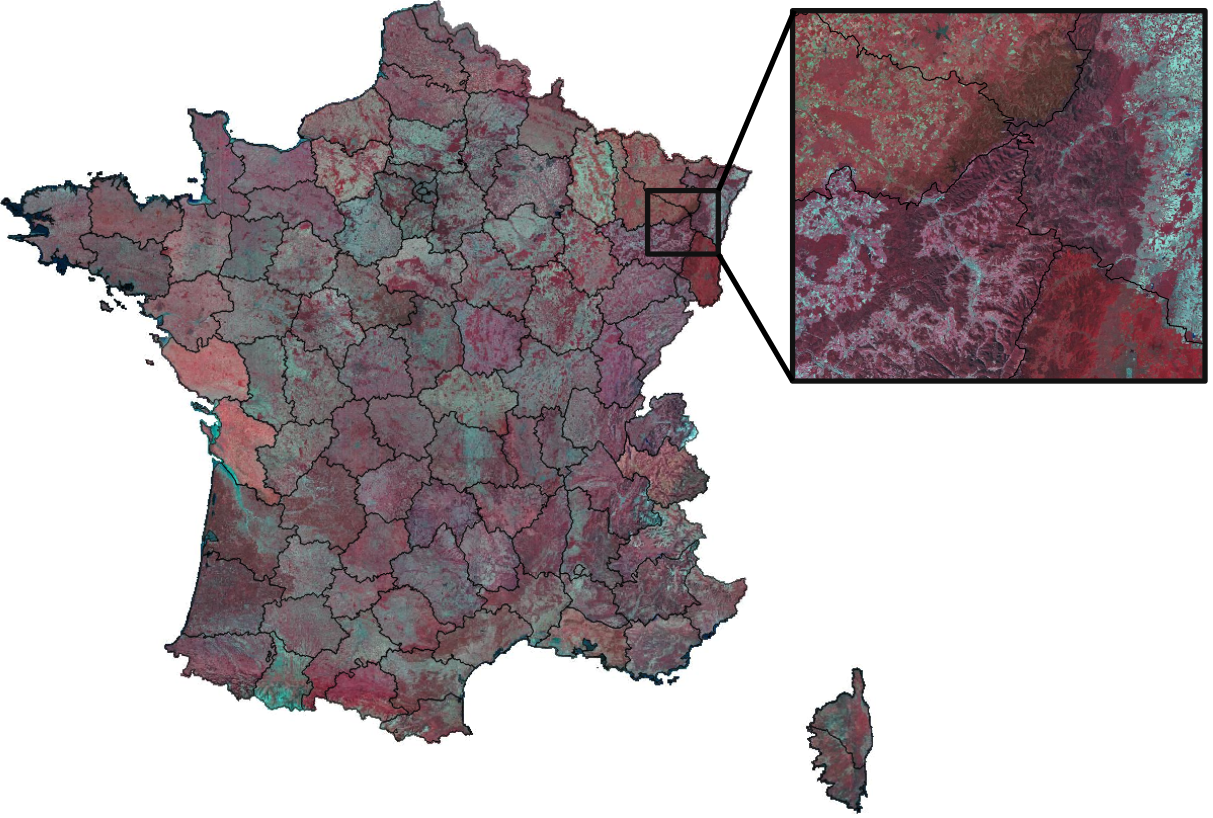}
  \caption{Radiometric discrepancies of the aerial images between domains. The bands displayed are composite of near-infrared, red and green spectral information. Figure adapted from \cite{flair_paper}.}
  \label{fig:radio}
\end{figure}

For this reason, we experiment a new lightweight Convolutional Neural Network (CNN), named GeoMultiTaskNet (GeoMTNet), on a new dataset (FLAIR i.e., French Land cover from Aerospace ImageRy \cite{flair_paper}), properly shaped for UDA (see for example the radiometric shifts in Fig. \ref{fig:radio}).
This contribution is the first in which the FLAIR dataset is used for scientific purposes.

GeoMTNet is a novel algorithm for UDA in semantic segmentation of RS images leveraging geographical coordinates, to align the source and target domains, with two key novelties.
First, we propose a simple GeoMultiTask module (GeoMT) that learns to predict the geographic position of the input image.
Second, inspired by \cite{li2022unsupervised}, we propose a Dynamic Class Sampling (DCS) module that adapts the semantic segmentation loss to the frequency of the classes. 

To our knowledge, this is the first work to address UDA in semantic segmentation using geographical metadata. The proposed approach offers a promising solution for reducing the annotation cost in semantic segmentation of VHR RS images, with a simple and portable module. Our proposed method establishes on a subset of the FLAIR dataset new state-of-the-art performance (47.22\% mIoU) with a limited number of parameters (33M), w.r.t. the transformer counterparts (85M).

\section{Related Work}
\label{sec:relworks}

\subsection{Unsupervised Domain Adaptation}

UDA approaches could be divided into three main branches: feature alignment, labeling adjustment, and discriminative methods. 
Feature alignment methods have the aim of aligning some characteristics (e.g., color histograms or features) of source and target domains. Some examples are DeepCORAL \cite{sun2016deep}, KeepItSimple \cite{abramov2020keep}, CoVi \cite{na2022contrastive}, GtA \cite{wang2021domain}. 
Labeling adjustment makes use of pseudo-labeling to force the predictions of the target domain to be consistent. Several works followed these strategies, such as NoisyStudent \cite{xie2020self}, CBST \cite{zou2018unsupervised}.
Discriminative methods are based on loss terms that force the net to distinguish among source and target features, e.g., DANN \cite{ganin2016domain}, AdaptSegNet \cite{Tsai_adaptseg_2018}, ADVENT \cite{vu2019advent}, DADA \cite{vu2019dada}.
Moreover, some hybrid approaches are also developed. For example, we can recall methods based on a combination of the presented strategies, such as SePiCo \cite{xie2022sepico}, DISE \cite{chang2019all} and DAFormer \cite{Hoyer_2022_CVPR, hoyer2022hrda}. Finally, hybrid UDA approaches such as self-supervised learning (SSL) \cite{wang2021domain, pan2020unsupervised, pipa} or continual learning \cite{saporta2022multi} have been explored.

\subsection{UDA for Remote Sensing}
Different methods, not all aiming properly for UDA, have been proposed. 
StandardGAN \cite{Tasar_2020_CVPR_Workshops} works with multi-source domains, forcing the domains to have similar distributions. Seasonal Contrast (SeCo) \cite{manas2021seasonal} is based on two steps: gathering uncurated RS images, then, using SSL. 
Bidirectional sample-class alignment (BSCA) \cite{HUANG2023192} addresses semi-supervised domain adaption for cross-domain scene classification. 
ConSecutive Pre-Training (CSPT) \cite{conspretr}, similarly to \cite{manas2021seasonal}, aims to leverage knowledge from unlabeled data through a self-supervision approach. 
MemoryAdaptNet \cite{zhu2022unsupervised} constructs an output space adversarial learning framework to tackle domain shift.
UDAT \cite{ye2022unsupervised} addresses UDA for nighttime aerial tracking, through a transformer. 
MATerial and TExture Representation Learning (MATTER) \cite{Akiva_2022_CVPR} aligns domains of different datasets, through a self-supervision task, on several tasks.
UDA\_for\_RS \cite{li2022unsupervised}, complementing \cite{Hoyer_2022_CVPR}, proposes a Gradual Class Weights (GCW) and a Local Dynamic Quality (LDQ) module.

\subsection{Using Geographical Metadata}
The first attempts at using geoinformation, outside the UDA framework, were presented in \cite{liao2015tag, tang2015improving}.
In \cite{mai2022review}, the authors provide a comprehensive review of location encoding. \cite{mac2019presence} proposes an efficient spatiotemporal prior, that estimates the probability that a given object category occurs at a specific location. GeoKR \cite{GeoKR} uses metadata for an efficient pre-training strategy on a wide dataset.
In \cite{Baudoux}, geographical coordinates are used for map translation. Geography-Aware SSL \cite{ayush2021geography} proposes an SSL algorithm based on the geoinformation of the patches.
In \cite{mai2020multi}, the authors present Space2Vec to encode the absolute positions and location spatial relationships. PE-GNN \cite{graphgeo} follows a similar approach, using graphs.

\section{Methodology}
\label{sec:method}

\begin{figure*}
  \centering
  \includegraphics[width=0.75\linewidth]{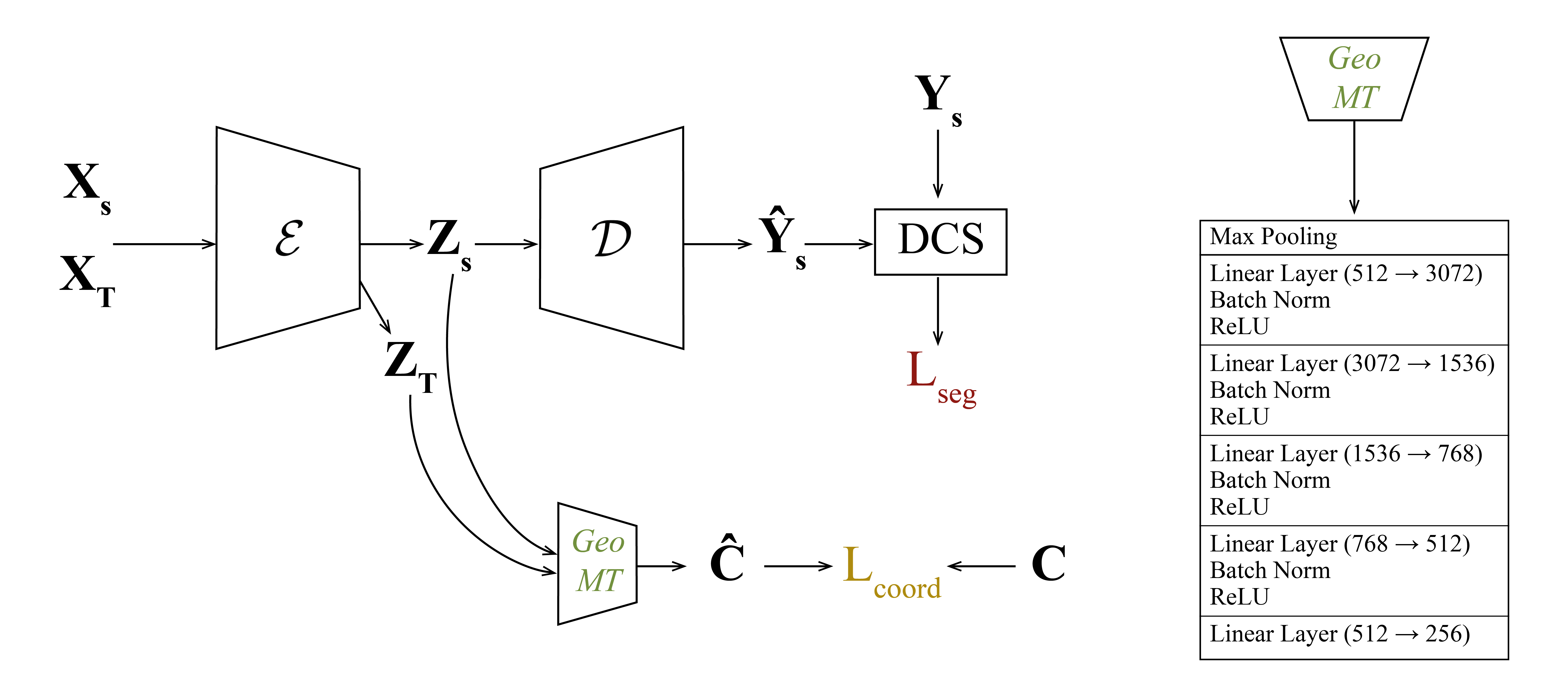}
  \caption{On the left the overview of the proposed architecture, made of: an encoder ($\mathcal{E}$), a decoder ($\mathcal{D}$), the GeoMultiTask module, and the Dynamic Class Sampling module. On the right, the structure of the GeoMultiTask module with input and output sizes.}
  \label{fig:arch}
\end{figure*}

As stated, the aim of GeoMTNet is to reduce domain shift using geographic coordinates, by designing a lightweight and easy-to-use architecture. In particular, given a set of source images $\mathbf{X}_S \in \mathbb{R}^{H \times W \times B}$, where $H$ is the height of the images, $W$ is the width and $B$ are the bands of the images in input, and a set of target images $\mathbf{X}_T \in \mathbb{R}^{H \times W \times B}$, we want to predict the annotation maps of the target, $\hat{\mathbf{Y}}_T \in \mathbb{R}^{H \times W}$, making use only of $\mathbf{X}_S$, $\mathbf{X}_T$ and the labels of the source images, $\mathbf{Y}_S \in \mathbb{R}^{H \times W}$. The labels of the target images, $\mathbf{Y}_T \in \mathbb{R}^{H \times W}$, could only be used for evaluation purposes. To achieve these targets, we decided to adopt a classic U-Net \cite{ronneberger2015u} as the backbone model.
It is easy to train and commonly used. It has a reduced number of parameters w.r.t. transformer counterparts. It exploits a semantic segmentation training, through the use of a pixel level-classification loss. On the other hand, to tackle domain shift, visible in Fig. \ref{fig:radio}, the GeoMultiTask (GeoMT) module is added. The net is further trained with the Dynamic Class Sampling (DCS) strategy.
Both of them have been shaped to be easily portable to different architectures. First, the GeoMT makes use of the geographical coordinates as a proxy to supervise the domain shift. Inspired by different self-supervised approaches \cite{ayush2021geography, manas2021seasonal}, we consider that an effective method to improve the performance on the target domains is to constrain, through a loss term, the features of the encoder to understand where the target images are located. Considering that also the source domain is made of different sub-domains (i.e. departments), the GeoMT is employed to constrain the encoder to learn generalized representations of all the data. Second, as the distribution of the labels is skewed, we propose DCS, to limit the errors on the under-represented classes, inspired by \cite{li2022unsupervised}. The whole architecture is shown in Fig. \ref{fig:arch}. In the next sections, the GeoMultiTask module (Section \ref{sec:geomt}) and the Dynamic Class Sampling (Section \ref{sec:dcs}) are presented in a formal and detailed way.

\subsection{GeoMultiTask Module}
\label{sec:geomt}

In other EO tasks, some approaches used geographical coordinates, such as using them as residual \cite{Baudoux} or skip connections, or even being stacked to the input \cite{yang2021semantic}. 
In our case, inspired by SSL \cite{manas2021seasonal, ayush2021geography}, we decided to use coordinates to drive and constrain the encoder features. Specifically, both $\mathbf{X}_S$ and $\mathbf{X}_T$ images pass through the encoder $\mathcal{E}$. This results in $\mathbf{Z}_S \in \mathbb{R}^{H' \times W' \times C}$ and $\mathbf{Z}_T \in \mathbb{R}^{H' \times W' \times C}$ feature maps, where $H', W'$ and $C$ are the height, width and number of channels of the feature maps. $\mathbf{Z}_S$ passes through the decoder $\mathcal{D}$ to obtain $\hat{\mathbf{Y}}_S$. 
In parallel, both representations $\mathbf{Z}_S$ and $\mathbf{Z}_T$ enter the GeoMT to predict a vector $\widehat{\mathbf{C}} \in \mathbb{R}^D$ containing localization information. Specifically, each patch is assigned a pair of coordinates $(C_{lon}, C_{lat})$, referring to the centroid of the patch itself. These coordinates undergo the following transformations to be used as supervision for $\widehat{\mathbf{C}}$:
\begin{enumerate}
    \item centering them in the reference system EPSG:2154, w.r.t. whom the coordinates are expressed. Particularly, we subtract $x = 489353.59\:m$ to $C_{lon}$ and $y = 6587552.20\:m$ to $C_{lat}$, to make the median values equal $(0,0)$;
    \item noise injection of $30\:km$ to let the net capture large-scale patterns not too specifically referred to the patches in the batch, but rather to wider areas of France, that may eventually even cross the boundaries of individual departments;
    \item positional encoding of the coordinates, for similar reasons of the noise injection. The strategy uses the following formula:
\begin{equation}
\mathbf{C}=\left[\begin{array}{c}
\sin \left(C_{lon} \omega_1\right) \\
\cos \left(C_{lon} \omega_1\right) \\
\vdots \\
\sin \left(C_{lat} \omega_{d / 4}\right) \\
\cos \left(C_{lat} \omega_{d / 4}\right)
\end{array}\right]_d \text { with } \omega_i=\frac{1}{f^{2 i / d}},
\end{equation}
where $D=256$ and $f=20,000$. Particularly, for the same reasons of the noise injection, $f$ is set to 20,000 and not 10,000 like in most applications \cite{vaswani2017attention, Baudoux}.
\end{enumerate}

GeoMT consists, firstly, of a max-pooling layer, which is used to reduce dimensionality and select the most meaningful features. After this, 5 linear layers, 4 of which employ batch normalization and ReLUs, are stacked. The detailed sizes are given in right part of Fig. \ref{fig:arch}. This module produces $\widehat{\mathbf{C}}$ from which we compute the self-supervised loss $L_{coord}$ that has the form of a mean squared error:

\begin{equation}
L_{coord} = {\frac{1}{n}\sum_{i=1}^{n}\left(\widehat{\mathbf{C}}_{i} - \mathbf{C}_{i}\right)^{2}},
\end{equation}
where $n$ is the number of samples.

The final loss of the GeoMTNet is thus:
\begin{equation}
L = L_{seg} + L^{S}_{coord} + L^{T}_{coord},
\end{equation}
where $L_{seg}$ is the segmentation loss, computed among $\widehat{\mathbf{Y}}_S$ and $\mathbf{Y}_S$, $L^{S}_{coord}$ is the loss term referred to the source domain images, and $L^{T}_{coord}$ to the target ones.

\subsection{Dynamic Class Sampling}
\label{sec:dcs}

Class imbalance is a common problem in deep learning, that leads to poor model generalization, especially in rare classes. To address this issue, researchers have proposed various methods, such as assigning class weights inversely proportional to the frequency of the class in the dataset \cite{zou2018unsupervised}. The class weight for class $c$, referred to the $n$-th label, is calculated as follows:

\begin{equation}
    w(n, c)=\frac{N_{c} \cdot \exp \left[\left(1-f_c\right) / t\right]}{\sum_{c^{\prime}=1}^C \exp \left[\left(1-f_{c^{\prime}}\right) / t\right]},
\end{equation}

where $f_c$ is the frequency of class c in the training dataset, $N_{c}$ is the total number of classes, and $t$ is a temperature parameter. The frequency $f_c$ is calculated as:

\begin{equation}
    f_c=\frac{1}{H \times W} \sum_{h=1}^H \sum_{w=1}^W\left(y_S^{(h, w)}\right)_c,
\end{equation}

where $y_S^{(h, w)}$ denotes the one-hot source label at location $(h,w)$ in the image, and $(\cdot)_c$ denotes the $c$-th scalar of a vector. Inspired by \cite{li2022unsupervised}, which applies a similar mechanism to the pseudo-labels predicted by the student network, the class weight is updated iteratively for each image using an exponentially weighted average:
\begin{equation}
    \operatorname{DCS}(n, c)=\alpha \cdot \operatorname{DCS}(n-1, c)+(1-\alpha) \cdot w(n, c).
\end{equation}

$\alpha$ is the decay rate of the exponential average. It helps to reduce volatility, especially in the early stages of training.
Unlike other approaches \cite{li2022unsupervised}, this weighting strategy  does not impact the pseudo-labels but the predicted labels directly.
The distribution of the classes will be different from the whole dataset in advance, due to sampling randomness: the weights will be updated iteratively for each image.
It is also worth noting that, instead of directly initializing the class weights to the distributions estimated from the first sample, they are initialized to 1 and then updated iteratively by an exponentially weighted average. A higher $t$ leads to a more uniform distribution. A lower one makes the model pay more attention to the rare classes. 

The final segmentation loss is:

\begin{equation}
    L_{seg}=-\sum_{h=1}^{H} \sum_{w=1}^{W} \operatorname{DCS}(n, c) \cdot y_{S}^{(h,w)} \cdot \log \left(h_{\theta}\left(x_{S}^{(h,w)}\right)\right),
\end{equation}

where $h_{\theta}$ is the model with weights $\theta$.

\section{Dataset}
\label{sec:data}

The French National Institute of Geographical and Forest Information (IGN) \cite{ign} is a French public state administrative establishment in charge of measuring large-scale changes on the French territory. It is constructing the French national reference land cover map \textit{Occupation du sol à grande  échelle} (OCS-GE), also making use of AI-based data and techniques. To this purpose, IGN developed the FLAIR dataset\footnote{downloadable at \url{https://ignf.github.io/FLAIR/}}.

\subsection{FLAIR dataset}
\label{sec:dataset}

\begin{figure*}
  \centering
  \includegraphics[width=1\linewidth]{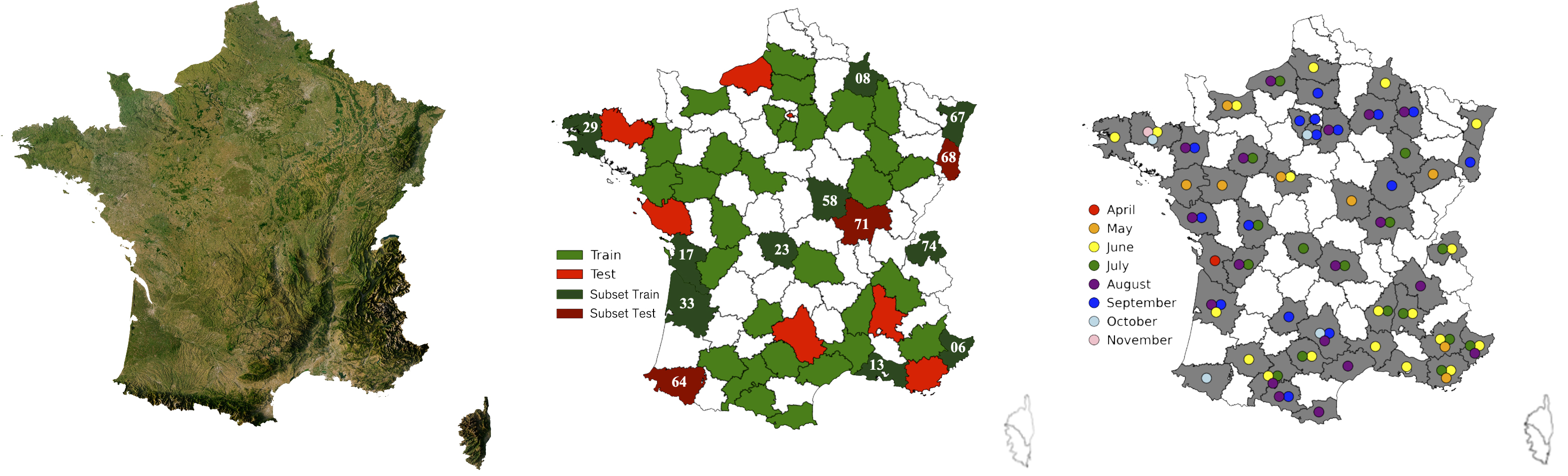}
  \caption{On the left, the ORTHO HR® aerial image cover of France. On the center, the train and test split of the 50 domains, with the domains selected for our experiments highlighted. On the right, the acquisition time of each domain. Figure adapted from \cite{flair_paper}.}
  \label{fig:france}
\end{figure*}

The French Land cover from Aerospace ImageRy (FLAIR) dataset \cite{flair_paper} includes 50 spatial domains varying along the different landscapes and climates of metropolitan France. Each domain is a French department (Fig. \ref{fig:france}).

The complete dataset is composed of 77,412 patches, covering approximately $810\:km^2$. Each patch is $512\times512$ pixels, with a ground sample distance (GSD) of $0.2 \:m$. Each domain is composed of $1725-1800$ patches. The domains were selected considering the major landscapes (e.g., urban, agricultural, etc..) and per semantic class radiometries (see Fig. \ref{fig:radio}).
To acquire the images, more than three years were needed. This led to a high intra- and inter-domain variance in the acquisitions (see Fig. \ref{fig:france} and \ref{fig:radio}).
The images have 5 bands corresponding to blue, green, red, near-infrared and elevation channels. The first 4 channels are retrieved from VHR aerial images ORTHO HR® \cite{ortohr}. The fifth channel is obtained through the difference between the Digital Surface Model and the Digital Terrain Model (see \cite{flair_paper} for more details). The corresponding ground truth labels describe the semantic class for each pixel. Nineteen classes are annotated. The \textit{other} class corresponds to pixels impossible to define with certainty. Finally, the dataset is split into 40 domains for the training and 10 for testing, ensuring a comparable distribution of the labels in train and test. The domains are highlighted in Fig.~\ref{fig:france}.
Each patch is enriched with metadata:
\begin{itemize}
    \item domain and zone label. The zone label is made of two letters, allowing a macro-distinction of the two major types of land cover of the area. The letter U indicates urban, N natural area, A agricultural area, and F forest. 
    \item date and hour at which the aerial image was acquired;
    \item the geographical coordinates of the centroid and the mean altitude of the patch;
    \item camera type used during aerial image acquisition \cite{souchon2014}.
\end{itemize}
To our knowledge, this is the first time that this dataset has been used for scientific research. 
Particularly, in our experiments, we used a subset of the whole dataset: D06, D08, D13, D17, D23, D29, D33, D58, D67, D74 as source domains and D64, D68, D71 as target domains. We ended up with more than 16k images for training and more than 5k for testing. 

\section{Experimental Setup}
\label{sec:expset}

As stated, for our experiments, we selected 10 departments as source domains and 3 as target domains. Adopting the same strategy as \cite{flair_paper}, we considered as \textit{other} all the classes labeled as $>12$. These classes are strongly under-represented, being $<0.2\%$ of all the labels.
Thus, we ended up with 13 classes (i.e. \textit{building}, \textit{pervious surface}, \textit{impervious surface}, \textit{bare soil}, \textit{water}, \textit{coniferous}, \textit{deciduous}, \textit{brushwood}, \textit{vine}, \textit{grassland}, \textit{crop}, \textit{plowed land}, \textit{other}).
A single Tesla V100-SXM2 32 GB GPU was used for the training phase. Having limited computational power, but still wanting to preserve the high resolution of the dataset (GSD = $0.2\:m$), for the training, we used random crops of $256 \times 256$. For the testing stage, we perform inference on four non-overlapping crops of $256 \times 256$, for each patch of size $512 \times 512$. For the U-Net, we use a ResNet18 \cite{he2016deep}, pretrained on ImageNet, as encoder and the softmax function as activation on the last layer. For all the experiments, we fix the batch size to 16, the number of epochs to 120 and the learning rate to 0.0001. We used early stopping with a patience of 30 epochs. The semantic segmentation loss is a cross-entropy, ignoring the \textit{other} class. We used Adam as optimizer and RandAugment \cite{cubuk2020randaugment} as the set of augmentations. The mean intersection over union (mIoU) on the first 12 classes is the selected evaluation metric.
For the DCS module, we set the parameters to $T=0.9$ and $\alpha = 0.7$.
To assess the performance of our strategy, we selected different methods\footnote{We tested them through the code in their official GitHub repositories.} from the literature for an extensive comparison. We chose: AdaptSegNet \cite{Tsai_adaptseg_2018}, which employs an adversarial training approach; ADVENT \cite{vu2019advent}, using an entropy minimization strategy; DAFormer \cite{Hoyer_2022_CVPR}, which adopts a transformer with a self-training strategy and UDA\_for\_RS \cite{li2022unsupervised}, that optimizes the DAFormer for RS tasks.
In Section \ref{sec:expres}, we present different experimental results, addressing both comparisons and several ablation studies.

\section{Experimental Results}
\label{sec:expres}

GeoMTNet reaches satisfying performance, shown in Tables \ref{tab:comp} and \ref{tab:full_comp}.  As expected, the under-represented classes, such as \textit{coniferous} and \textit{brushwood}, are the most difficult to be correctly predicted. This is due both to the few quantities of data and the radiometric similarity with some more frequent classes. For example, \textit{coniferous} could be easily confused with \textit{deciduous}. At the same time, some errors are due to the fact that images share some similar spatial patterns. This is the case of \textit{vine} and \textit{crop} pixels. Another frequent misclassification error concerns \textit{bare soil}. Even though the performance is satisfying (55\% mIoU), we can see that the variance implicit in the definition of this class led to confusion with \textit{herbaceous cover} or \textit{impervious surface}. 
In the next sections, comparisons (Section \ref{sec:comp}) and ablation studies (Sections \ref{sec:gmt}, \ref{sec:gtmt} and \ref{sec:mod}) are carried one.

\subsection{Comparison}
\label{sec:comp}

\begin{figure*}
  \centering
  \includegraphics[width=0.85\linewidth]{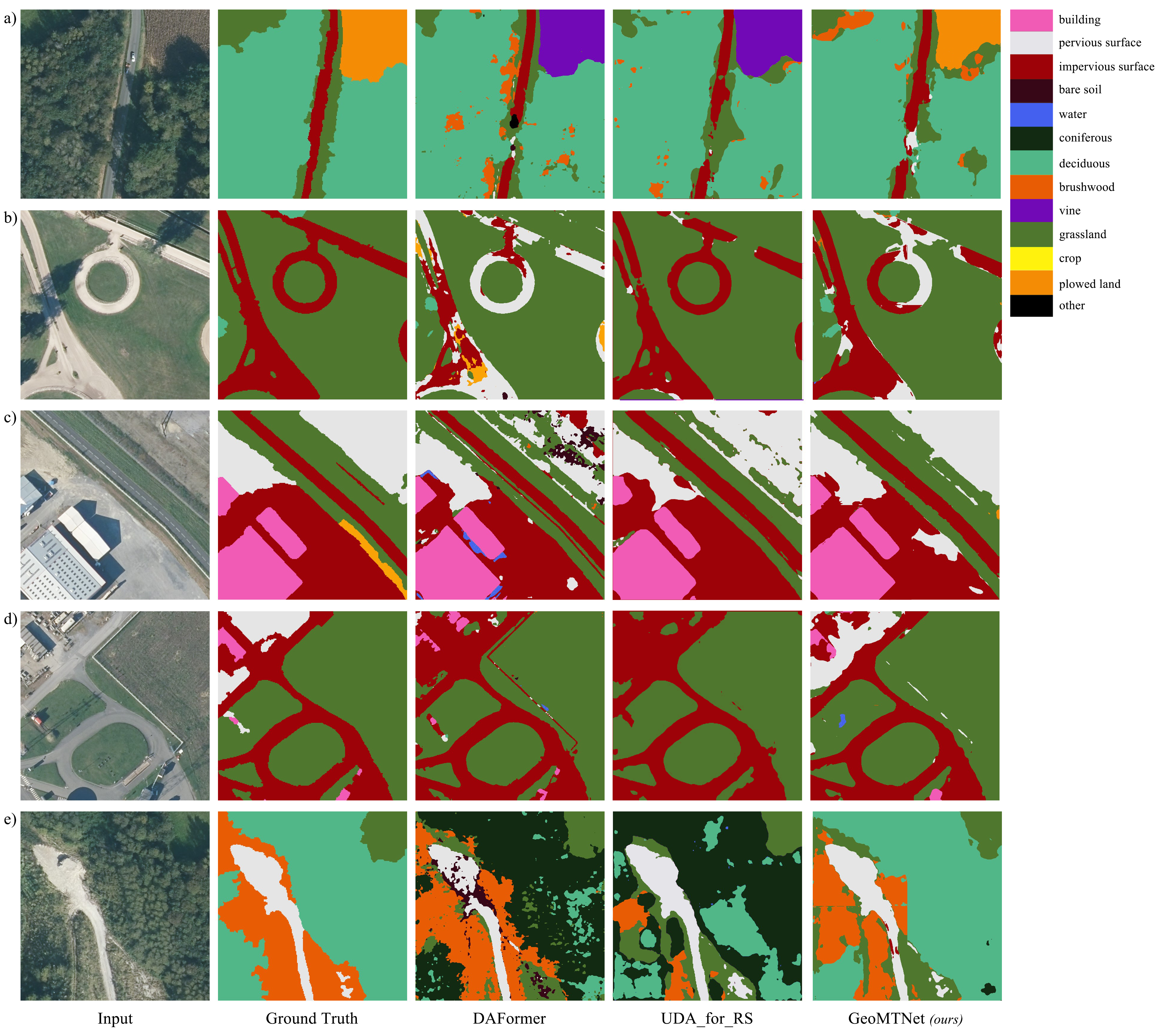}\\
  \caption{Some examples of predictions for the best performing models. Particularly we can see in order: the input image, the ground truth, the prediction of DAFormer, the prediction of UDA\_for\_RS and the prediction of our GeoMTNet.}
  \label{fig:comp}
\end{figure*}

Despite using a smaller number of parameters (33M), GeoMTNet reaches better results than all the other selected architectures (47.22\% mIoU). In particular, we can see from Table \ref{tab:comp} that there is a deep gap w.r.t. AdaptSegNet \cite{Tsai_adaptseg_2018} (24.97\% mIoU) and ADVENT \cite{vu2019advent} (25.56\% mIoU), which are more dated and, probably, properly developed for the synthetic-to-real benchmarks \cite{gtadata, synthiadata}. 
On the other hand, DAFormer \cite{Hoyer_2022_CVPR} and UDA\_for\_RS \cite{li2022unsupervised}, based on a transformer, have comparable performance with the GeoMTNet (respectively 45.61\% and 47.02\% mIoU). When using strategies properly shaped for RS task, such as in UDA\_for\_RS \cite{li2022unsupervised}, optimal results are obtained. However, from both Table \ref{tab:comp}, Table \ref{tab:full_comp} and Fig. \ref{fig:comp}, we can see that GeoMTNet leads to better results also w.r.t. the aforementioned method with a reduced number of parameters (33M for GeoMTNet vs 85M for UDA\_for\_RS). 
Focusing on the detailed performance, reported in Table \ref{tab:full_comp}, we can observe that GeoMTNet almost has the best performance on all the classes, except for four of them (that are \textit{pervious surface}, \textit{bare soil}, \textit{brushwood}, and \textit{vine}). This is mainly justified by the fact that each different architecture tends, when deciding among two similar classes, to overestimate one of them and underestimate the other. For example, \textit{vine} is often confused with \textit{plowed land} (and sometimes \textit{crop}, too), due to their similar pattern. DAFormer, still having a gain of more than 10\% in IoU over GeoMTNet performance for \textit{vine}, reaches poor results both on \textit{plowed land} (41.83\% vs 54.79\% in IoU) and \textit{crop} (23.74\% vs 35.02\% in IoU). This phenomenon could be observed also in Fig. \ref{fig:comp}, where some predictions of the three best models (namely DAFormer, UDA\_for\_RS, and GeoMTNet) are reported to draw some qualitative results. 
We observe that DAFormer performs overall worse, as it often predicts some irrelevant classes, with a poorer texture and shape of the polygons predicted. 
On the other hand, most of the time UDA\_for\_RS predicts a smaller number of classes with a wider predicted area for each of them w.r.t. the other methods. This is mainly due to the LDQ module of UDA\_for\_RS, which bases the pixel prediction on the predictions made on the contiguous pixels. This can be seen both in positive cases, Fig. \ref{fig:comp} b), where the land cover prediction of the traffic circle is more consistent than in GeoMTNet, and in negative cases, Fig. \ref{fig:comp} d), where the low confidence in predicting \textit{pervious surface} and \textit{building} ends in a uniformed incorrect prediction of \textit{impervious surface}. On the other hand, we can appreciate the consistency in shape reconstructions and boundaries in GeoMTNet more than in the others (see, for example, the building edges in Fig. \ref{fig:comp} c)). 
Moreover, we can see how shadows consist in an important problem for all the architectures (see for example in Fig. \ref{fig:comp} a) how the shape of the \textit{plowed land} in the upper right part of the image is badly reconstructed for both UDA\_for\_RS and GeoMTNet).
Another issue to consider is that train patches are of size $512\times512$ while the model is trained on $256\times256$ patches. Thus, sometimes, the borders of the predicted tiles to have contrasting predictions, as visible in the central part of Fig. \ref{fig:comp} e).

\begin{table}[h]
\centering
\begin{tabular}{ccc}
Architecture                    & mIoU (\%)      & params (M)  \\ \hline
AdaptSegNet \cite{Tsai_adaptseg_2018} & 24.97          & 99          \\
ADVENT    \cite{vu2019advent}                      & 25.56           & 99          \\
DAFormer    \cite{Hoyer_2022_CVPR}                    & 45.61          & 85          \\
UDA\_for\_RS     \cite{li2022unsupervised}               & 47.02          & 85          \\
GeoMultiTaskNet \textbf{(ours)} & \textbf{47.22} & \textbf{33}
\end{tabular}
\caption{Our GeoMultiTaskNet outperforms all the other methods on the considered FLAIR target domains. In addition to the improved results in terms of mIoU, the size of the proposed model is also significantly smaller than that of the other selected algorithms.}
\label{tab:comp}
\end{table}

\begin{table*}[h]
\adjustbox{max width=\textwidth}{\begin{tabular}{ccccccccccccc}
\multirow{2}{*}{Architecture} &
  \multicolumn{12}{c}{IoU (\%)} \\ \cline{2-13} 
 &
  building &
  pervious surface &
  impervious surface &
  bare soil &
  water &
  coniferous &
  deciduous &
  brushwood &
  vine &
  grassland &
  crop &
  plowed land \\ \hline
AdaptSegNet \cite{Tsai_adaptseg_2018} &
  39.98 &
  20.75 &
  40.23 &
  20.36 &
  15.25 &
  4.93 &
  35.37 &
  10.99 &
  34.51 &
  42.69 &
  11.06 &
  23.47 \\
ADVENT    \cite{vu2019advent} &
  35.79 &
  24.38 &
  48.82 &
  6.85 &
  31.98 &
  0.00 &
  51.65 &
  11.79 &
  33.33 &
  25.76 &
  11.46 &
  24.29 \\
DAFormer    \cite{Hoyer_2022_CVPR} &
  67.09 &
  45.56 &
  61.99 &
  55.35 &
  65.12 &
  8.91 &
  54.39 &
  \textbf{20.31} &
  \textbf{64.39} &
  38.79 &
  23.74 &
  41.83 \\
UDA\_for\_RS     \cite{li2022unsupervised}  &
  66.3 &
  \textbf{48.05} &
  62.36 &
  \textbf{59.28} &
  61.24 &
  9.22 &
  60.02 &
  16.52 &
  57.74 &
  40.12 &
  30.32 &
  54.17 \\
GeoMultiTaskNet \textbf{(ours)} &
  \textbf{67.53} &
  40.86 &
  \textbf{63.89} &
  55.31 &
  \textbf{67.02} &
  \textbf{13.85} &
  \textbf{60.97} &
  14.08 &
  53.09 &
  \textbf{40.33} &
  \textbf{35.02} &
  \textbf{54.79}
\end{tabular}
}
\caption{Comparison in the IoU for each class of the considered FLAIR target domains.}
\label{tab:full_comp}
\end{table*}

\subsection{GeoMultiTask module}
\label{sec:gmt}

To understand the GeoMTNet capabilities, various informative ablation studies have been conducted. To perform these experiments in an easy and rapid way, we used a simple U-Net \cite{ronneberger2015u} as CNN, with less than 2M parameters. As mentioned before, the GeoMT takes as input the features provided by the encoder and tries to infer low-frequency encoded coordinates, with a random noise injection. As it could be argued from Tables \ref{tab:geonoiseab} and \ref{tab:geosizeab}, both of these strategies improve the net performance.

At first, we focused on the challenge of using coordinates, still feeding the GeoMT with the decoder features.
As stated, the goal is to allow the net to capture large-scale patterns, not too specific to the single patch, but rather to areas of France that may eventually even cross the boundaries of individual departments. We tried two strategies: positional encoding and noise injection.
Positional encoding\cite{vaswani2017attention}, used in EO approaches \cite{Baudoux}, allows the coordinates to be represented with a vector, making it easier to grasp reciprocal phenomena of proximity between patches. Noise injection allows to make net performance more generalizable, avoiding the association of a specific coordinate with a specific patch. In light of these considerations, we have tried different configurations (Table \ref{tab:geonoiseab}).
Notably, using a lower frequency (i.e., 1/20000) than the one used in the literature \cite{vaswani2017attention, Baudoux}, brings greater benefits. In fact, we are interested in large-scale effects, enhanced by a lower frequency. Concerning noise injection, it has been empirically demonstrated that a consistent noise ($30\:km$) compared to the size of a patch (about $100\:m$) helps the generalization process. However, increasing it too much (about $50\:km$) leads to excessive network confusion and a consequent drop in performance.

Secondly, we needed to limit the number of parameters, especially w.r.t. the other models in the literature. To do this, we have no longer used as input of the GeoMT the features in output from the decoder, but those in output from the encoder. In fact, the encoder features should already provide the necessary information to perform a correct segmentation.
This intuition was supported by the results, shown in Table \ref{tab:geosizeab}, which shows a slight drop in performance, completely negligible. Notably, the shape of the GeoMT is also slightly different. In the case described so far (input features of the decoder), the module consists of two convolutional layers (to reduce the dimensionality of the features with a limited number of parameters) followed by three linear layers.

\begin{table}[]
\centering
\begin{tabular}{cccc}
noise (km) & 1/frequency (-) & mIoU (\%) & params (M) \\ \hline
- & - & 42.05 & 1.9 \\
- & - & 41.69 & 270 \\
- & 10,000 & 43.57 & 270 \\
±30 & 10,000 & 43.69 & 270 \\
±30 & 20,000 & 44.83 & 270 \\
±50 & 20,000 & 42.68 & 270 
\end{tabular}
\caption{Ablation studies, showing the behavior of GeoMultiTaskModule under different noise injections and encoding.}
\label{tab:geonoiseab}
\end{table}

\begin{table}[]
\centering
\begin{tabular}{ccc}
input features & mIoU (\%) & params (M) \\ \hline
- & 42.05 & 1.9 \\
output of the decoder & 44.83 & 270 \\
output of the encoder & 44.70 & 11.2
\end{tabular}
\caption{Ablation studies, showing the behavior of GeoMultiTaskModule when using different input features.}
\label{tab:geosizeab}
\end{table}

\subsection{GeoTimeMultiTask experiments: using the temporal information}
\label{sec:gtmt}

We tried to include temporal information as well, also inspired by other works \cite{manas2021seasonal, ayush2021geography}. In particular, we inserted both month and time of day information, discarding the year. In fact, the month impacts the seasonality of some classes (e.g., the vegetative ones), and the hour the acquisition conditions \cite{cong2022satmae}. 
As previously, we tried to inject some noise, so that the features could generalize better. Specifically, the time information was circle encoded (i.e., arranged equally spaced on a circle) and, when used, a random noise of $\pm1$ was added. Finally, these experiments were carried out using either the encoder or the decoder features. In both circumstances, the TimeMultiTask module (TimeMT) has been defined similarly to the GeoMT, but smaller in size. For example, in the case of using the encoder features, the TimeMT consists of one max-pooling layer and two linear layers. 
We refer to these experiments as GeoTimeMT, being characterized by both GeoMT and TimeMT modules. Also for this set of experiments, a simple U-Net was used as the backbone, without ResNet as the encoder. The results are shown in Table \ref{tab:timeab}. Two behaviors can be observed immediately: using temporal metadata leads to limited improvements (+1.72\% mIoU w.r.t. the baseline); GeoTimeMT, which combines geographical and temporal information, does not improve results obtained using only GeoMT (43.77\% vs 44.70\% mIoU). For these reasons, our GeoMTNet makes only use of geographical coordinates.

\begin{table}
\centering
\resizebox{\columnwidth}{!}{\begin{tabular}{ccccc}
input features & \multicolumn{1}{l}{time used} & \multicolumn{1}{l}{time noise} & mIoU (\%) & params (M) \\ \hline
- & - & - & 42.05 & 1.9 \\
decoder & both & no & 38.19 & 405 \\
encoder & both & no & 42.72 & 11.9 \\
encoder & month & yes & 43.77 & 11.9
\end{tabular}}
\caption{Ablation studies, showing the behavior of GeoTimeMultiTaskModule under different conditions.}
\label{tab:timeab}
\end{table}

Analyzing the detailed results, we can observe that using the features outputted by the encoder is more beneficial than the ones outputted by the decoder, both from a performance (38.19\% vs 42.72\% mIoU) and size point of view (405M vs 11.9M parameters). In fact, the benefits derived from temporal metadata are more related to the features directly encoded from the images, such as radiometric information of the images, more than to the decoded representations of the patches, such as the one connected to land cover. In addition, we observe again that using less precise information, thus with noise injection, leads to better results (42.72\% vs 43.77\% mIoU). Finally, we observe that the hour information is less relevant than the month information. In fact, the large variance of the dataset and the large amount of images, make it more important and beneficial to have representations from different seasons of same classes more than under different light conditions. In fact, classes such as \textit{brushwood} or \textit{crop} really vary their radiometric information depending on the seasonality.

\subsection{Comprehensive baselines}
\label{sec:mod}

We considered important to evaluate the impact of each component of the GeoMTNet. The results of these experiments are shown in Table \ref{tab:bas18}.

\begin{table}[h]
\centering
\begin{tabular}{ccc}
net                          & mIoU (\%) & params (M) \\ \hline
baseline                     & 42.51     & 25         \\
+GeoMT                       & 46.68     & 32.7       \\
+DCS                         & 43.25     & 25         \\
\multicolumn{1}{l}{GeoMTNet} & \textbf{47.22}     & 32.7      
\end{tabular}
\caption{Ablation studies about the component of the GeoMultiTaskNet. As stated, both components lead to better results than the baseline, even though the GeoMultiTask module performs better.}
\label{tab:bas18}
\end{table}

The component that leads to the greatest improvement is the GeoMT which leads to a gain of about 4\% in mIoU, while DCS does not go beyond a percentage point. This is due to the fact that GeoMT is properly shaped to enhance RS metadata, to empower the architecture on which it is applied. 
In contrast, unlike other approaches such as \cite{li2022unsupervised}, in our GeoMTNet, the weighting module, namely DCS, does not impact the pseudo-labels, but the predicted labels directly. Therefore, its effectiveness on images from target domains is influenced directly by the source images.

\section{Conclusions}
\label{sec:conc}

In light of major technological innovations, more and more RS images are available. However, the annotation of these images is not progressing at the same rate, leading to a vast amount of unlabelled data. Most of the time, these images carry metadata, which are often simply discarded for CV tasks. In this work, we showed that the use of architectures specifically designed to exploit such metadata in an EO context can lead to excellent results. To this end, we proposed GeoMultiTaskNet, which outperforms other models in the literature, despite being a lightweight network, on the FLAIR dataset. This real-world scenario-oriented dataset presents a great variety of information and is well-suited for this type of experiments. In this context, this work only presents itself as a first step in a line of research that is as important as it is still under-investigated: remote sensing unsupervised domain adaptation. Future steps include the extension of GeoMultiTaskNet over the entire FLAIR dataset. In addition, the intention is to probe this model on other datasets \cite{xia2023openearthmap}, where domain shift is more important, and to find new ways to integrate geo-metadata into already performing models, such as transformers.

\small
\bibliographystyle{ieee_fullname}
\bibliography{PaperForReview}

\end{document}